\documentclass[10pt,twocolumn,letterpaper]{article}

\usepackage{wacv}
\usepackage{times}
\usepackage{epsfig}
\usepackage{graphicx}
\usepackage{amsmath}
\usepackage{amssymb}
\usepackage{url}
\usepackage{tabu}
\usepackage{xcolor}
\usepackage{multirow,tabularx}
\usepackage{array}
\usepackage{makecell}
\usepackage{tikz, pgfplots} \usetikzlibrary{positioning} \tikzset{basic/.style={draw,fill=blue!20,textwidth=1em,text badly centered}} \tikzset{input/.style={basic,circle}} \tikzset{weights/.style={basic,rectangle}} \tikzset{functions/.style={basic,circle,fill=blue!10}} \tikzset{activation/.style={basic,circle}}

\wacvfinalcopy 

\def\wacvPaperID{****} 

\ifwacvfinal\fi
\begin{document}

\title{Structural Recurrent Neural Network (SRNN) for Group Activity Analysis}

\author{Sovan Biswas \\
University of Bonn\\
{\tt\small sbiswas@uni-bonn.de}
\and
Juergen Gall \\
University of Bonn\\
{\tt\small gall@iai.uni-bonn.de}
}

\maketitle
\ifwacvfinal\thispagestyle{empty}\fi

\begin{abstract}
A group of persons can be analyzed at various semantic levels such as individual actions, their interactions, and the activity of the entire group.  
In this paper, we propose a structural recurrent neural network (SRNN) that uses a series of interconnected RNNs to jointly capture the actions of individuals, their interactions, as well as the group activity. While previous structural recurrent neural networks assumed that the number of nodes and edges is constant, we use a grid pooling layer to address the fact that the number of individuals in a group can vary.      
We evaluate two variants of the structural recurrent neural network on the Volleyball Dataset.

\end{abstract}

\section{Introduction}
\label{sec:intro}

Activity analysis has been of great interest in computer vision since decades. In recent years, deep learning approaches such as \cite{carreira2017kinnetics,ji2013cnn3d,simonyan2014twostream,singh2014birnn,karpathy2014video} have been proposed to recognize activities in videos. Most of these approaches, however, focus on single person activity analysis and estimate only one activity per video clip. Similar to other recent works \cite{ramanathan2016eventactor,deng2016sim,alahi2016sociallstm,shu2017cern}, the goal of this paper is to understand and analyze the actions of individuals and their interactions, and subsequently use them for predicting the group activity. 

Recent deep learning approaches for group activity analysis such as \cite{msibrahiCVPR16deepactivity,shu2017cern} use a multi-level hierarchy of recurrent neural networks (RNNs) for group activity recognition. In these approaches, the lower level RNNs focus on understanding and modeling the actions of individuals and the higher level RNNs in the architecture model the group activity. These approaches are trained using a two-step process. The first step focuses on improving the recognition of the actions of each individual independently and the subsequent step focuses on recognizing the group activity given the recognized actions of the individuals.

\begin{figure}[t!]
 \centering
  \includegraphics[width=0.4\textwidth]{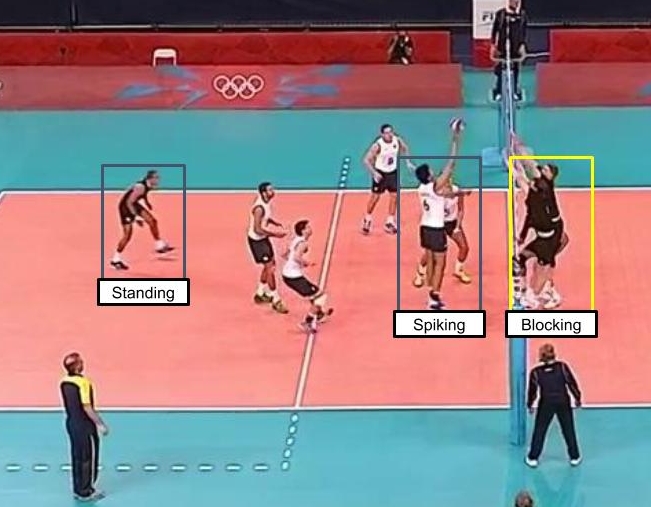}
 \caption{A frame labeled as group activity ``Left Spike" and bounding boxes around each team player are annotated in the dataset with individual actions~\cite{msibrahiPAMI16deepactivity}.} 
 \label{fig:left_spike}
 \end{figure}
 
Apart from the hindrance of two-step training, these methods lack the capability of capturing interactions between individual persons when present within a group. 
For example, in a volleyball game as shown in Figure \ref{fig:left_spike}, a person from the team on the left-hand side of the volleyball court performs the individual action ``spiking" whereas the player from the opponent team performs a ``blocking" action. Looking only at the individuals makes it difficult to distinguish between the individual actions, but there is a strong correlation between the two activities. Similarly, when walking in a crowd, people move and walk in various directions just to avoid colliding with each other. In general, the action of an individual in a group is influenced by the actions of the other individuals in the group. 
This phenomenon not only provides context that helps to recognize the individual actions but also provides a key information about the group level actions such as in the case of volleyball as shown in Figure \ref{fig:left_spike}. Thus, there is an imperative need to analyze interactions between individuals and to capture the influence over time when analyzing a group of humans. 

The main focus of the paper is to harness such interactions within a group to improve the recognition of the group activity as well as the individual actions. To this end, we build on the recently proposed structural recurrent neural network (SRNN) \cite{jain2016srnn} which has the unique capability of capturing interactions as contextual information using an interconnected set of RNNs. While in \cite{jain2016srnn}, the number of nodes and edges and therefore the number of RNNs is constant, we extend the approach to handle a varying number of nodes and edges as it is required for analyzing group activities.    

   

The rest of the paper is structured as follows. We start with a brief discussion of the related work in Section \ref{sec:reltdwrk}, followed by a short introduction of SRNNs in Section \ref{sec:srnndef}. In Section \ref{sec:algo}, we introduce two variants of an SRNN. We then evaluate the two variants in Section \ref{sec:exp} and conclude with a summary in Section \ref{sec:summary}.

\section{Related Work}
\label{sec:reltdwrk}
One of the earlier approaches for analyzing the activities of a group or crowd was proposed by \cite{choi2009they}. They introduce crowd context in recognizing the activity being performed by each individual in the group. Traditionally graphical models with key contextual features \cite{choi2014collectiveactivity,choi2011context} have been deployed rigorously towards group analysis. However these models with handcrafted features \cite{ryoo09} were outperformed by newer deep neural network architectures such as \cite{deng2016sim,alahi2016sociallstm,shu2017cern}. For group activity recognition, most of these deep neural networks are inspired by \cite{msibrahiCVPR16deepactivity} which uses a multi-level cascade of recurrent neural networks for group activity recognition. In this approach, humans are detected and tracked to form multi-person tracklets. These tracklets along with their deep visual features are fed to the lower level RNNs. The focus of these lower level RNNs is to understand and model the actions of the individual persons. The higher level RNNs in the architecture instead focus on understanding the group activity. The individual actions and group activity predictions are done using softmax in a feed-forward way. However, each method tackles a very different problem in the same framework. Shu \etal \cite{shu2017cern} use a similar hierarchical architecture but the approach differs from previous work by proposing an energy based approach that works significantly better if the amount of data is small. 
Furthermore, this approach also explores human interaction, but holistically by convolutional features extracted from both humans. \cite{bagautdinov2017socialscene} propose a joint approach for detecting humans and predicting their actions.   

Spatio-temporal graphs have been used in computer vision for various applications such as predicting human movements \cite{ionescu2014mocap} or learning human activities and object affordances \cite{koppula2013affordances}. The spatio-temporal graphs represent in these works spatio-temporal relations between joints or joints and objects in a video.  
The methods \cite{koppula2016anticipate,amer2014hirf} use handcrafted features along with graphical models, conditional random field or random forest for the aforementioned applications. Recently, neural networks have been deployed to solve spatio-temporal graph problems. For example, \cite{deng2015structured} uses deep networks followed by inference using a probabilistic graphical model to recognize the actions in a group. Our approach builds on the work \cite{jain2016srnn} where a set of coupled RNNs are used to represent spatio-temporal graphs.    

\section{Structural RNN}
\label{sec:srnndef}
Recurrent neural networks are very effective in modeling temporal sequences. In case of a single person, features $f^t$ are extracted at each frame and used as input for an RNN to predict the action classes $y^t$ over time.  
However when in a group, a person performs an action based on its interaction with other persons and the group objective. So, a single recurrent neural network is incapable of capturing the interactions and group dynamics, thus reducing its effectiveness. For solving similar problems, Jain \etal \cite{jain2016srnn} proposed an interconnected set of recurrent neural networks that not only captures the individual behavior over time but also integrates the interactions between the individuals through edges. 
 
 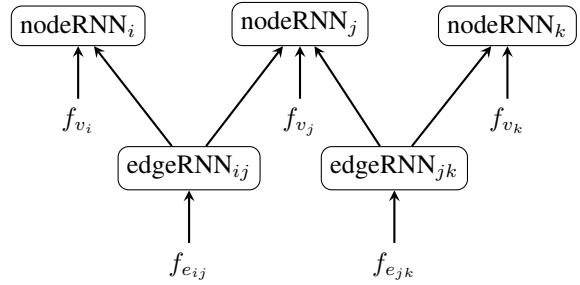
\begin{figure}[t]
\begin{minipage}{0.48\textwidth}
 \centering
 \begin{tikzpicture}[node distance=1cm]
\tikzstyle{io} = [rectangle, rounded corners, minimum width=1cm, minimum height=0.5cm,text centered, draw=black]
\tikzstyle{layer} = [rectangle, rounded corners, minimum width=1.5cm, minimum height=0.5cm,text centered, draw=black]
\tikzstyle{biglayer} = [rectangle, rounded corners, minimum width=5cm, minimum height=0.5cm,text centered, draw=black]
\tikzstyle{midlayer} = [rectangle, rounded corners, minimum width=2.5cm, minimum height=0.5cm,text centered, draw=black]
\tikzstyle{translayer} = [rectangle, rounded corners, minimum width=0.2cm, minimum height=0.5cm,text centered, draw=none]
\tikzstyle{arrow} = [thick,->,>=stealth]

\node (n0) [layer] {edgeRNN$_{ij}$};
\node (n1) [layer, right=0.8cm of n0] {edgeRNN$_{jk}$};
\node (n3) [left=0.4cm of n0] {};
\node (n4) [right=0.4cm of n0] {};
\node (n5) [right=0.4cm of n1] {};

\node (l0) [translayer, below=0.7cm of n0] {$f_{e_{ij}}$};
\node (l1) [translayer, below=0.7cm of n1] {$f_{e_{jk}}$};

\draw [arrow] (l0) to (n0);
\draw [arrow] (l1) to (n1);

\node (o0) [layer, above=1.5cm of n3] {nodeRNN$_{i}$};
\node (o1) [layer, above=1.5cm of n4] {nodeRNN$_{j}$};
\node (o2) [layer, above=1.5cm of n5] {nodeRNN$_{k}$};

\draw [arrow] (n0) to (o0);
\draw [arrow] (n0) to (o1);
\draw [arrow] (n1) to (o1);
\draw [arrow] (n1) to (o2);

\node (b0) [translayer, below=0.7cm of o0] {$f_{v_{i}}$};
\node (b1) [translayer, below=0.7cm of o1] {$f_{v_{j}}$};
\node (b2) [translayer, below=0.7cm of o2] {$f_{v_{k}}$};

\draw [arrow] (b0) to (o0);
\draw [arrow] (b1) to (o1);
\draw [arrow] (b2) to (o2);

\end{tikzpicture} 
 \caption{Feedforward network of a Structural RNN (SRNN) when trained with respect to node labels.}
 \label{fig:srnn}
 \end{minipage}
 \end{figure}
 
An example of an SRNN is illustrated in Figure~\ref{fig:srnn}. It consists of three nodes $v_i$, $v_j$, and $v_k$ and the goal is to predict for each node the class labels over time, which are denoted by $y_{v_i}^t$,  $y_{v_j}^t$, and  $y_{v_k}^t$, respectively. Each node is modeled by an RNN, termed nodeRNN. It takes as input some features $f_{v_i}^t$, which are extracted for a node $v_i$, but also the output of RNNs that model the interactions with other nodes. The second type of RNN is termed edgeRNN. The edgeRNNs take as input some features $f_{e_{ij}}^t$ based on the spatio-temporal relation between two nodes $v_i$ and $v_j$ and predicts a latent representation $h_{e_{ij}}^t$, which is forwarded to the corresponding nodeRNNs. The advantage of such an SRNN is that the nodeRNNs and the edgeRNNs can be trained jointly such that the prediction of the node labels depends not only on the features that are extracted for each node but also on the interactions between the nodes.

\section{Group Activity Analysis}
\label{sec:algo}
In this section, we will first briefly introduce the problem in Section~\ref{ssec:graph}. This is followed by introducing two different variants of an SRNN for group activity recognition in Section \ref{ssec:algo}.
 

\subsection{Problem Formulation}
\label{ssec:graph}

Our objective is to predict jointly the group activity label $y^t_g$ of a group as well as the action label $y^t_{v_i}$ for each individual $v_i$ of the group over time. We assume that the bounding box for each person has been already extracted and we compute features for each individual person $f^t_{v_i}$ and for each edge $f^t_{e_{ij}}$ between two individuals. The features are described in Section \ref{ssec:imple} and we denote the set of all node and edge features for a frame by $F^t$. In order to learn the parameters $\theta$ of the models which are described in Section~\ref{ssec:algo}, we minimize the loss  
\begin{align}
	\nonumber\underset{\theta}{\arg\min} \text{ } \Big[&L\left(\psi_{g}\left(F^t;\theta\right),y_g^t\right) \\
	&+ \frac{1}{n}\sum_{i=1}^n{L\left(\psi_{v}\left(F^t;\theta\right),y^t_{v_i}\right)} \Big],
	\label{eq:loss}
\end{align}
where $L$ denotes the cross-entropy loss, $\psi_{g}(.)$ the prediction function of the model for the group activity, and $\psi_{v}(.)$ the prediction function for the actions of the individuals.

\subsection{SRNN for Group Activity Analysis}
\label{ssec:algo}

\begin{figure*}[t!]
\begin{minipage}{0.45\textwidth}
	 \centering
  \begin{tikzpicture}[node distance=1cm]
\tikzstyle{io} = [rectangle, rounded corners, minimum width=1cm, minimum height=0.5cm,text centered, draw=black]
\tikzstyle{layer} = [rectangle, rounded corners, minimum width=1.5cm, minimum height=0.5cm,text centered, draw=black]
\tikzstyle{biglayer} = [rectangle, rounded corners, minimum width=5cm, minimum height=0.5cm,text centered, draw=black]
\tikzstyle{midlayer} = [rectangle, rounded corners, minimum width=2.5cm, minimum height=0.5cm,text centered, draw=black]
\tikzstyle{translayer} = [rectangle, rounded corners, minimum width=0.2cm, minimum height=0.5cm,text centered, draw=none]
\tikzstyle{arrow} = [thick,->,>=stealth]

\node (n0) [layer] {edgeRNN$_{ij}$};
\node (n1) [layer, right=0.7cm of n0] {edgeRNN$_{jk}$};
\node (n2) [layer, right=0.7cm of n1] {edgeRNN$_{ki}$};
\node (n3) [right=0.35cm of n0] {};
\node (n4) [right=0.35cm of n1] {};
\node (n5) [right=0.35cm of n2] {};

\node (l0) [translayer, below=0.7cm of n0] {$f_{e_{ij}}$};
\node (l1) [translayer, below=0.7cm of n1] {$f_{e_{jk}}$};
\node (l2) [translayer, below=0.7cm of n2] {$f_{e_{ki}}$};

\node (m0) [biglayer, above=0.7cm of n1] {Grid Pooling};

\draw [arrow] (l0) to (n0);
\draw [arrow] (l1) to (n1);
\draw [arrow] (l2) to (n2);

\draw [arrow] (n0) to (m0);
\draw [arrow] (n1) to (m0);
\draw [arrow] (n2) to (m0);

\node (o0) [layer, above=2.5cm of n0] {nodeRNN$_{i}$};
\node (o1) [layer, above=2.5cm of n1] {nodeRNN$_{j}$};
\node (o2) [layer, above=2.5cm of n2] {nodeRNN$_{k}$};

\node (b0) [translayer, below=0.5cm of o0] {};
\node (b00)[translayer, left=0.0cm of b0] {$f_{v_{i}}$};
\node (b1) [translayer, below=0.5cm of o1] {};
\node (b11) [translayer, left=0.0cm of b1] {$f_{v_{j}}$};
\node (b2) [translayer, below=0.5cm of o2] {};
\node (b22) [translayer, right=0.0cm of b2] {$f_{v_{k}}$};

\node (a0) [translayer, above=0.6cm of o0] {};
\node (a00)[translayer, left=0.0cm of a0] {$y_{v_{i}}$};
\node (a1) [translayer, above=0.6cm of o1] {};
\node (a11) [translayer, left=0.0cm of a1] {$y_{v_{j}}$};
\node (a2) [translayer, above=0.6cm of o2] {};
\node (a22) [translayer, right=0.0cm of a2] {$y_{v_{k}}$}; 

\draw [arrow] (m0) to (o1);
\draw [arrow] (m0) to (o0);
\draw [arrow] (m0) to (o2);

\draw [arrow] ([xshift=-0.485cm]o0.north) to (a00);
\draw [arrow] ([xshift=-0.485cm]o1.north) to (a11);
\draw [arrow] ([xshift=0.485cm]o2.north) to (a22);

\draw [arrow] (b00) to ([xshift=-0.485cm]o0.south);
\draw [arrow] (b11) to ([xshift=-0.485cm]o1.south);
\draw [arrow] (b22) to ([xshift=0.485cm]o2.south);

\node (p0) [biglayer, above=1.5cm of o1] {Max Pooling};

\draw [arrow] (o0) to (p0);
\draw [arrow] (o1) to (p0);
\draw [arrow] (o2) to (p0);

\node (q0) [midlayer, above=0.7cm of p0] {Group RNN};

\draw [arrow] (p0) to (q0);

\node (f0) [translayer, above=0.6cm of q0] {$y_g$};
\draw [arrow] (q0) to (f0);

\end{tikzpicture}
 \centering
 \caption{SRNN-MaxNode: Feedforward SRNN where max pooling is performed over the nodeRNNs. The nodeRNNs are enriched using the output of the edgeRNNs using a novel grid pooling approach.} 
 \label{fig:strat1}
\end{minipage}
\hspace{0.1cm}
\begin{minipage}{0.55\textwidth}
	 \centering
\begin{tikzpicture}[node distance=1cm]
\tikzstyle{io} = [rectangle, rounded corners, minimum width=1cm, minimum height=0.5cm,text centered, draw=black]
\tikzstyle{layer} = [rectangle, rounded corners, minimum width=1.5cm, minimum height=0.5cm,text centered, draw=black]
\tikzstyle{biglayer} = [rectangle, rounded corners, minimum width=5cm, minimum height=0.5cm,text centered, draw=black]
\tikzstyle{midlayer} = [rectangle, rounded corners, minimum width=2.5cm, minimum height=0.5cm,text centered, draw=black]
\tikzstyle{translayer} = [rectangle, rounded corners, minimum width=1.5cm, minimum height=0.5cm,text centered, draw=none]
\tikzstyle{arrow} = [thick,->,>=stealth]

\node (n0) [layer] {nodeRNN$_i$};
\node (n1) [layer, right=0.7cm of n0] {nodeRNN$_j$};
\node (n2) [layer, right=0.7cm of n1] {nodeRNN$_k$};
\node (n3) [right=0.35cm of n0] {};
\node (n4) [right=0.35cm of n1] {};
\node (n5) [right=0.35cm of n2] {};
\node (n6) [layer, right=0.7cm of n2] {nodeRNN$_i$};

\node (l0) [translayer, below=0.7cm of n0] {$f_{v_{i}}$};
\node (l01) [translayer, below=0.7cm of n3] {$f_{e_{ij}}$};
\node (l1) [translayer, below=0.7cm of n1] {$f_{v_{j}}$};
\node (l12) [translayer, below=0.7cm of n4] {$f_{e_{jk}}$};
\node (l2) [translayer, below=0.7cm of n2] {$f_{v_{k}}$};
\node (l20) [translayer, below=0.7cm of n5] {$f_{e_{ki}}$};
\node (l3) [translayer, below=0.7cm of n6] {$f_{v_{i}}$};

\node (a0) [translayer, above=0.6cm of n0] {};
\node (a00)[translayer, right=-2.0cm of a0] {$y_{v_{i}}$};
\node (a1) [translayer, above=0.6cm of n1] {$y_{v_{j}}$};
\node (a2) [translayer, above=0.6cm of n2] {$y_{v_{k}}$};
\node (a3) [translayer, above=0.6cm of n6] {};
\node (a22) [translayer, right=-1.cm of a3] {$y_{v_{i}}$};

\draw [arrow] ([xshift=-0.47cm]n0.north) to (a00);
\draw [arrow] (n1) to (a1);
\draw [arrow] (n2) to (a2);
\draw [arrow] ([xshift=0.5cm]n6.north) to (a22);

\node (n01) [layer, above=2.5cm of n3] {edgeRNN$_{ij}$};
\node (n12) [layer, above=2.5cm of n4] {edgeRNN$_{jk}$};
\node (n20) [layer, above=2.5cm of n5] {edgeRNN$_{ki}$};

\draw [arrow] (l0) to (n0);
\draw [arrow] (l01) to (n01);
\draw [arrow] (l1) to (n1);
\draw [arrow] (l12) to (n12);
\draw [arrow] (l2) to (n2);
\draw [arrow] (l20) to (n20);
\draw [arrow] (l3) to (n6);

\draw [arrow] (n0) to (n01);
\draw [arrow] (n1) to (n01);

\draw [arrow] (n1) to (n12);
\draw [arrow] (n2) to (n12);

\draw [arrow] (n2) to (n20);
\draw [arrow] (n6) to (n20);

\node (m0) [biglayer, above=1.5cm of n12] {Max Pooling};

\draw [arrow] (n01) to (m0);
\draw [arrow] (n12) to (m0);
\draw [arrow] (n20) to (m0);

\node (o0) [midlayer, above=0.7cm of m0] {Group RNN};

\draw [arrow] (m0) to (o0);

\node (f0) [translayer, above=0.6cm of o0] {$y_g$};
\draw [arrow] (o0) to (f0);

\end{tikzpicture}
	\centering 
 \caption{SRNN-MaxEdge: Feedforward SRNN where max pooling is performed over the edgeRNNs. The edgeRNNs are enriched using the output of the nodeRNNs.}
 \label{fig:strat2}
\end{minipage}

\end{figure*}

The proposed group activity recognition approach is formulated as a two-level hierarchy of recurrent neural networks similar to \cite{msibrahiCVPR16deepactivity,shu2017cern}. The lower level predicts individual actions followed by the higher level recurrent network that estimates the group activity. In Sections~\ref{sec:maxnode} and \ref{sec:maxedge}, we discuss two SRNN variants that jointly estimate the group activity and the individual actions by modeling the interactions between individuals. The two SRNNs are shown in Figures \ref{fig:strat1} and \ref{fig:strat2}.

\subsubsection{SRNN-MaxNode}\label{sec:maxnode}
As shown in Figure \ref{fig:left_spike}, when a person on the left hand side jumps to perform the action ``spiking", the opponents jump to block the spike, resulting in the action ``blocking" across the volleyball net. This is an example where persons in a group perform contextual actions. Structural RNNs are an efficient approach to model such relations. 

\begin{figure}[t]
 \centering
  \includegraphics[width=0.4\textwidth]{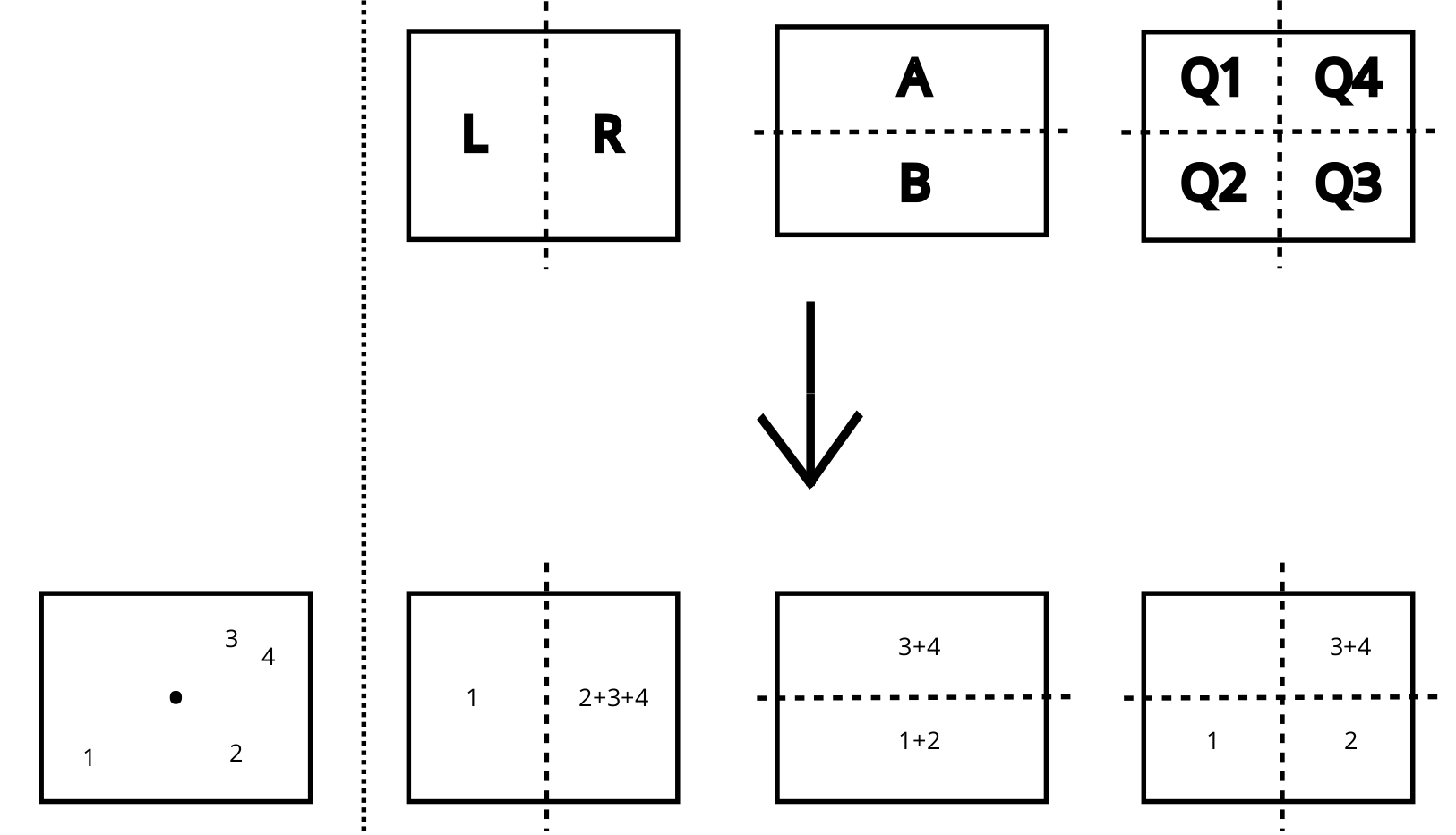}
 \caption{Grid pooling. Left: $1$, $2$, $3$ and $4$ denote four persons in the neighborhood of the person $\bullet$. Right: We define three grid structures (top) and we sum the outputs of the edgeRNNs where the neighbors of $\bullet$ are in the same cell. We then concatenate the features of the eight cells. If a cell is empty, the feature vector is set to zero.}
 \label{fig:grid_pool}
 \end{figure}

As discussed in Section~\ref{sec:srnndef}, SRNNs are a hierarchy of RNNs consisting of edgeRNNs and nodeRNNs. The first variant that we propose for group activity analysis is shown in Figure \ref{fig:strat1}. The lowest level consists of edgeRNNs that model interactions between two individuals based on their relative position, which is encoded by the feature vector $f^t_{e_{ij}}$.   
The output of the edgeRNNs is feedforwarded to the nodeRNNs. The number of individuals, however, varies in a group and each individual might have a different number of neighbors and in very dense crowds the number of neighbors could be very high. The approach proposed in \cite{jain2016srnn} cannot handle such cases since it concatenates the features, assuming that the number of edges and nodes is constant.   
To address this problem, we propose a grid pooling layer that combines for a node $v_i$ the output from all edgeRNNs $e_{ij}$ based on the position of the neighboring persons $v_j$ in a prescribed grid. The prescribed grid regions are arranged as shown in Figure \ref{fig:grid_pool}.
If several neighbors are in the same grid cell, we sum the output of the edgeRNNs instead of averaging them. This means that the values are usually larger when more persons are in a cell. 
The grid pooling provides for each node $v_i$ features for 8 cells that are then concatenated with additional CNN features $f^t_{v_i}$, which are extracted from the frame $t$ for each person. The output of the nodeRNNs is used in two ways. First, the action class $y^t_{v_i}$ of the person $v_i$ is predicted using an additional softmax layer. Second, the group activity is estimated similar to \cite{msibrahiCVPR16deepactivity} by max-pooling the outputs of the nodeRNNs of a group at a time instant $t$, which is then used as input for an RNN that predicts the group activity $y^t_g$ over time.

In summary, the SRNN-MaxNode model is defined as follows: 
\begin{eqnarray}
	h^t_{e_{ij}} &=& \text{RNN}_e(h^{t-1}_{e_{ij}},f^t_{e_{ij}}) \nonumber \\
	h^t_{C_i} &=& \sum_{j \in S_{C_i}} h^t_{e_{ij}}  \nonumber \\
	h^t_{e_i} &=& [h^t_{L^i} \ldots h^t_{Q_4^i}] \nonumber \\
	h^t_{v_i} &=& \text{RNN}_v(h^{t-1}_{v_{i}},h^t_{e_i}, f^t_{v_i}).
	\label{eq:strat1srnn}
\end{eqnarray}
While $h^t_{e_{ij}}$ denotes the output from the edgeRNN for the nodes $v_i$ and $v_j$ at frame $t$, ${S_{C_i}}$ denotes the set of neighboring nodes of $v_i$ in the cell $C_i \in \{L^i, R^i, A^i, B^i, Q_1^i, Q_2^i, Q_3^i, Q_4^i\}$, which are the grid regions for $v_i$ as shown in Figure \ref{fig:grid_pool}. The accumulated values for each cell $h^t_{C_i}$ are then concatenated to the vector $h^t_{e_i}$ and the output of the nodeRNN is denoted by $h^t_{v_i}$.

The full architecture can thus be defined as: 
\begin{eqnarray}
	h^t_{v_i} &=& \text{SRNN}(f^t_{v_i},f^t_{e_{ij}}) \nonumber \\
	y^t_{v_i} &=& \phi_v(h^t_{v_i},W_v) \nonumber \\
	h^t_p &=& \max{(h^t_{v_1} \ldots h^t_{v_n})} \nonumber \\
	h^t_{g} &=& \text{RNN}_g(h^{t-1}_{g},h^t_p) \nonumber \\
	y^t_g &=& \phi(h^t_g,W_g), 
	\end{eqnarray} 
where $h^t_{v_i}$ denotes the output after the SRNN \eqref{eq:strat1srnn} and $W_v$ denotes the weights used in the softmax function $\phi_v(.)$ to predict the individual actions $y^t_{v_i}$. $h^t_p$ denotes the max-pooled representation over the complete group and $h^t_g$ denotes the output of the group RNN, which is then used 
by a softmax function $\phi(.)$ with weights $W_g$ to predict the group activity $y^t_g$.

\subsubsection{SRNN-MaxEdge}\label{sec:maxedge}

While the SRNN in Figure \ref{fig:strat1} uses the edgeRNNs to provide contextual information for the nodeRNNs, we also compare it to an SRNN where the nodeRNNs are at the lowest level of the hierarchy. In this case, the max pooling is not performed over the nodeRNNs but over the edgeRNNs and an additional grid pooling is not required since each edge consists of two nodes. We denote the second variant, which is shown in Figure \ref{fig:strat2}, SRNN-MaxEdge.


For SRNN-MaxEdge, the lower level of the hierarchy consists of nodeRNNs that predict the individual actions based on the individual CNN features $f^t_{v_i}$. The output from the nodeRNNs is forwarded to the corresponding edgeRNNs, which also take the edge features $f^t_{e_{ij}}$ as input. 
%
The output of the edgeRNNs at a time instant $t$ is max pooled and then used as input for an RNN that predicts the group activity $y^t_g$ over time.
  
In summary, the SRNN-MaxEdge model is defined as follows: 
\begin{eqnarray}
	h^t_{v_i} &=& \text{RNN}_v(h^{t-1}_{v_{i}}, f^t_{v_{i}}) \nonumber \\
	y^t_{v_i} &=& \phi_v(h^t_{v_i},W_v) \nonumber \\
	h^t_{e_{ij}} &=& \text{RNN}_e(h^{t-1}_{e_{ij}},h^t_{v_i}, h^t_{v_j},f^t_{e_{ij}}),
	\label{eq:strat2srnn} 
\end{eqnarray}
where $h^t_{v_i}$ denotes the output of the nodeRNN for person $v_i$ at a given time $t$ and $W_v$ are the weights used in the softmax function $\phi_v(.)$ to predict the individual action $y^t_{v_i}$. $h^t_{e_{ij}}$ denotes the output of the edgeRNNs. 

The full architecture can thus be defined as: 
\begin{eqnarray}
	h^t_{e_{ij}} &=& \text{SRNN}(f^t_{v_i},f^t_{e_{ij}}) \nonumber \\
	h^t_p &=& \max{(h^t_{e_{12}} \ldots h^t_{e_{mn}})} \nonumber \\
	h^t_{g} &=& \text{RNN}_g(h^{t-1}_{g},h^t_p) \nonumber \\
	y^t_g &=& \phi(h^t_g,W_g), 
	\end{eqnarray}	 
where $h^t_{e_{ij}}$ denotes the output from the SRNN \eqref{eq:strat2srnn}. While $h^t_p$ denotes the max pooled representation over all edges in the group, $h^t_g$ denotes the output of the group RNN, which is then used 
by a softmax function $\phi(.)$ with weights $W_g$ to predict the group activity $y^t_g$.

 \begin{table*}
\begin{center}
{
\begin{tabular}{|c|c|c|}
\hline
\textbf{Method}&\textbf{Group Activity Accuracy}&\textbf{Individual Action Recognition Accuracy}\\
\hline
Hierarchical LSTM \cite{msibrahiPAMI16deepactivity} (1 group) & 70.3\% & - \\
\hline
Hierarchical LSTM V1 (1 group) & 68.37\% & \bf{76.32\%} \\
\hline
Hierarchical LSTM V2 (1 group) & 73.89\% & \bf{76.32\%} \\
\hline
Hierarchical LSTM V3 (1 group) & \bf{74.01\%} & 75.96\% \\
\hline\hline
Hierarchical LSTM \cite{msibrahiPAMI16deepactivity} (2 groups) & 81.9\% & - \\
\hline
Hierarchical LSTM V1 (2 groups) & 78.37\% & \bf{76.32\%} \\
\hline
Hierarchical LSTM V2 (2 groups) & 81.33\% & \bf{76.32\%} \\
\hline
Hierarchical LSTM V3 (2 groups) & \bf{83.12\%} & 75.96\% \\
\hline
\end{tabular}
}
\end{center}
\caption{Comparison of various variations of the Hierarchical LSTM~\cite{msibrahiPAMI16deepactivity} using Alexnet features.}
\label{tab:results3}
\end{table*}

\section{Experiments}
\label{sec:exp}
\subsection{Datasets}
We evaluate our framework on the recently introduced
volleyball dataset \cite{msibrahiPAMI16deepactivity}. This dataset has 55 volleyball game video sequences with 4830 labeled frames, where each player is labeled and subsequently annotated with the bounding box. Each player performs one of the 9 individual actions resulting in one of the 8 group activity labels. Furthermore, the whole dataset is divided into non-overlapping sets of 24 sequences for training, 15 sequences for validation and the remaining sequences are used for testing. Similar to \cite{msibrahiCVPR16deepactivity,shu2017cern}, we have used both training and validation sequences for training. Since not all frames are annotated by bounding boxes, the Dlib tracker \cite{king2009dblibtrack} is used to propagate the ground-truth bounding boxes to the unannotated frames. 


\subsection{Implementation Details}
\label{ssec:imple}
For the RNNs, we use standard LSTMs \cite{hochreiter1997lstm} and the implementation is done using the Tensorflow library. 
At the node level, each LSTM is connected with a deep convolutional network such as Alexnet \cite{krizhevsky2012alexnet} or VGG 16 \cite{Simonyan14c} to compute the  visual features $f_{v_i}$ based on the annotated and tracked bounding boxes of the persons. Similar to \cite{msibrahiCVPR16deepactivity,shu2017cern}, we initialize the CNNs by a model that has been pre-trained on ImageNet. During training, we fine-tune only the last two fully connected layers of the CNNs. 

The edge features $f^t_{e_{ij}}$ model spatio-temporal relations between the bounding boxes of two persons.      
We take the center of each bound box and compute the difference vector $(dx,dy)$. We then compute the basic distance values ($|dx|$, $|dy|$, $|dx+dy|$, $\sqrt{(dx)^2+(dy)^2}$) and add the direction of the translational vector ($arctan(dy, dx)$, $arctan2(dy, dx)$). To further enhance these simple 6 interaction features, we also compute the difference of the 6 features between two consecutive time frames, which results in 6 additional features. We finally compute the 12 features not only for frame $t$, but also for the neighboring frames $t-1$ and $t+1$ to capture some short temporal information. All features are concatenated to obtain a 36 dimensional feature vector.  

The training is performed in two stages. In the first stage, the nodeRNNs are trained independently using individual actions and the cross-entropy as loss function. For this stage, we have used a batch size of 36 for our experiments. In the next step, we train the whole architecture by minimizing the loss \eqref{eq:loss}. We use Adam as optimizer with a learning rate of 0.00001. During training, the parameters of the nodeRNNs and the last two layers of the CNN are updated as well.     
The nodeRNNs have 3000 hidden units and the group RNN has 2000 hidden units. The number of nodes for the edgeRNNs differs between the SRNN-MaxNode and the SRNN-MaxEdge due to differences of the input features. While the edgeRNNs in the SRNN-MaxNode take as input the low dimensional features $f^t_{e_{ij}}$, the edgeRNNs in the SRNN-MaxEdge use the additional output of two nodeRNNs as input. For the edgeRNNs in the SRNN-MaxNode, we use therefore only 30 hidden units whereas 1000 hidden units are used for the edgeRNNs in the SRNN-MaxEdge. Since the two variants differ in memory consumption and the GPU memory is limited, we use a batch size of 30 for SRNN-MaxNode and a batch size of 16 is used for SRNN-MaxEdge. 

In accordance with other approaches \cite{msibrahiCVPR16deepactivity,shu2017cern,bagautdinov2017socialscene}, we have also performed experiments where we divide the individuals in two groups. For this, we use the same approach as in Ibrahim \etal \cite{msibrahiPAMI16deepactivity}.

\begin{table*}
\begin{center}
{
\begin{tabular}{|c|c|c|}
\hline
\textbf{Method}&\textbf{Group Activity Accuracy}&\textbf{Individual Action Recognition Accuracy}\\
\hline
Hierarchical LSTM \cite{msibrahiPAMI16deepactivity} (1 group) & 70.3\% & - \\
\hline
Hierarchical LSTM V3 (1 group) & 74.01\% & 75.96\% \\
\hline
CERN \cite{shu2017cern} (1 group) & 73.5\% & 69\% \\
\hline
SRNN-MaxNode (1 group) & \textit{\bf{74.39\%}} & \textit{\bf{76.65\%}} \\
\hline
SRNN-MaxEdge (1 group) & 68.39\% & 76.03\% \\
\hline
\hline
Hierarchical LSTM\cite{msibrahiPAMI16deepactivity} (2 groups) & 81.9\% & - \\
\hline
Hierarchical LSTM V3 (2 groups) & 83.12\% & 75.96\% \\
\hline
CERN \cite{shu2017cern} (2 groups) & 83.3\% & 69\% \\
\hline
SRNN-MaxNode (2 groups) & \bf{83.47\%} & \bf{76.65\%} \\
\hline
SRNN-MaxEdge (2 groups) & 79.86\% & 76.03\% \\ \hline
\hline  
Social Scene \cite{bagautdinov2017socialscene} (2 groups) & \bf{89.9}\% & \bf{82.4}\% \\
\hline
\end{tabular}
}
\end{center}
\caption{Comparison to the state-of-the-art.}
\label{tab:results}
\end{table*}

\begin{table*}
\begin{center}
{
\begin{tabular}{|c|c|c|c|}
\hline
\textbf{Feature}&\textbf{Method}&\textbf{Group Activity Accuracy}&\textbf{Individual Action Recognition Accuracy}\\
\hline
\multirow{3}{*}{Alexnet} & H.\ LSTM V3 - (1 group) & 74.01\% & 75.96\% \\ 
\cline{2-4}
	& SRNN-MaxNode - (1 group) & \bf{74.39}\% & \bf{76.65}\% \\
	\cline{2-4}
	& SRNN-MaxEdge - (1 group) & 68.39\% & 76.03\% \\
\hline
\multirow{3}{*}{VGG 16} & H.\ LSTM V3 - (1 group) &  70.34\% & 75.30\% \\ 
\cline{2-4}
	& SRNN-MaxNode - (1 group) & 71.20\% & 74.85\% \\
	\cline{2-4}
	& SRNN-MaxEdge - (1 group) & 68.29\% & 75.96\% \\
	\hline
\hline
\multirow{3}{*}{Alexnet} & H.\ LSTM V3 - (2 groups) & 83.12\% & 75.96\% \\ 
\cline{2-4}
	& SRNN-MaxNode - (2 groups) & \bf{83.47}\% & \bf{76.65}\% \\
	\cline{2-4}
	& SRNN-MaxEdge - (2 groups) & 79.86\% & 76.03\% \\
\hline
\multirow{3}{*}{VGG 16} & H.\ LSTM V3 - (2 groups) &  81.34\% & 75.30\% \\ 
\cline{2-4}
	& SRNN-MaxNode - (2 groups) & 82.86\% & 74.85\% \\
	\cline{2-4}
	& SRNN-MaxEdge - (2 groups) & 79.92\% & 75.96\% \\
	\hline
\end{tabular}
}
\end{center}
\caption{Comparison of the proposed SRNN approaches with Hierarchical LSTM V3 using Alexnet or VGG 16 as CNN. }
\label{tab:results2}
\end{table*}

\subsection{Experimental Evaluation}
\label{sec:qres}
\subsubsection{Variation of Hierarchical LSTM}
\label{ssec:hlstm}
As the proposed approaches are inspired by Hierarchical LSTMs \cite{msibrahiPAMI16deepactivity}, we have first performed a few baseline experiments to evaluate versions of the Hierarchical LSTM \cite{msibrahiPAMI16deepactivity}: 
\begin{itemize}
	\item 2-layer LSTMs (V1): This is similar to \cite{msibrahiPAMI16deepactivity} with a two-step training for person level RNNs followed by training of the group level RNN given the pre-trained person RNNs. Unlike \cite{msibrahiCVPR16deepactivity}, the group level takes as input only the output of person RNNs and does not use any additional CNN features.   
	\item 2-layer LSTMs (V2): Reimplementation of \cite{msibrahiPAMI16deepactivity}.
	\item 2-layer LSTMs (V3): This is similar to V1 but person RNNs and group RNN are jointly trained using the loss \eqref{eq:loss}. As described in Section \ref{ssec:imple}, the last two layers of the Alexnet are fine-tuned.  
\end{itemize} 

The accuracy of recognizing the group activity as well as the actions of the individuals is reported in Table~\ref{tab:results3}. 
The results show that training the person RNNs and the group RNN jointly (V3) using the loss \eqref{eq:loss} improves the accuracy of the group activity also for the Hierarchical LSTM \cite{msibrahiPAMI16deepactivity}, but it slightly decreases the individual action recognition results. 
Dividing the individuals into two groups as in \cite{msibrahiPAMI16deepactivity} improves the accuracy by a large margin due to the volleyball scenario where two teams play against each other.

\subsubsection{Comparison to state-of-the-art}
\label{ssec:resss}
Table \ref{tab:results} compares the proposed SRNN approaches with the Hierarchical LSTM V3 that is trained with the same loss function and uses the same Alexnet CNN. While SRNN-MaxNode outperforms the Hierarchical LSTM both for group activity recognition as well as the recognition of the individual actions, SRNN-MaxEdge achieves a lower group activity accuracy than SRNN-MaxNode and Hierarchical LSTM. It shows that the max pooling over the nodeRNNs is better than pooling over the edgeRNNs on this dataset since the max pooling over the nodeRNNs forwards the features of the most important individual to the group RNN. This works for group activities as shown in Figure \ref{fig:left_spike} very well since the group activity can be well inferred from the ``spiking" person. Our proposed approach SRNN-MaxNode also outperforms the approach CERN \cite{shu2017cern}, which also uses Alexnet features. However, the recent approach \cite{bagautdinov2017socialscene}, which builds on the Inception-V3 CNN \cite{szegedy2016inceptionV3}, achieves the highest accuracy on this dataset.

%

 \subsubsection{Impact of CNN architecture}
We have also analyzed the impact of the CNN architecture and compare the used Alexnet CNN with the larger VGG 16 network. The results are reported in Table \ref{tab:results2}. The accuracy of the VGG 16 network decreases the accuracy of the Hierarchical LSTM as well as the proposed SRNN-MaxNode. For SRNN-MaxEdge the accuracy remains nearly the same. The decrease in accuracy might be due to overfitting, but it needs further investigation to analyze the impact of the used CNN model in more detail.         
 
%

\subsubsection{Impact of deep edge features}
For the model SRNN-MaxNode, we use a low dimensional feature vector $f^t_{e_{ij}}$ that encodes simple spatio-temporal relations between two bounding boxes. We also investigated if the accuracy can be improved when the edgeRNNs not only take $f^t_{e_{ij}}$ as input feature but also $f^t_{v_{i}}$ and $f^t_{v_{j}}$.     
Since this increases the dimensionality of the input feature from 36 to 8228 (4096+4096+36), we also increase the number of hidden units of the EdgeRNNs from 30 to 1000 to address the higher dimensionality. As shown in Table~\ref{tab:results4}, adding $f^t_{v_{i}}$ and $f^t_{v_{j}}$ does not improve the accuracy. This is expected since the features $f^t_{v_{i}}$ are already added to the nodeRNNs and adding them twice does not provide additional information for the model.    


\begin{table*}
\begin{center}
{
\begin{tabular}{|c|c|c|}
\hline
\textbf{Edge feature}&\textbf{Group Activity Accuracy}&\textbf{Individual Action Recognition Accuracy}\\
\hline
$f^t_{e_{ij}}$ (1 group) & 74.39\% & \textit{\bf{76.65\%}} \\
\hline
$(f^t_{e_{ij}}, f^t_{v_{i}}, f^t_{v_{j}})$ (1 group) & \textit{\bf{74.48\%}} & 75.89\% \\
\hline
\hline
$f^t_{e_{ij}}$ (2 groups) & \textit{\bf{83.47\%}} & \textit{\bf{76.65\%}} \\
\hline
$(f^t_{e_{ij}}, f^t_{v_{i}}, f^t_{v_{j}})$ (2 groups) & 83.27\% & 75.89\% \\
\hline
\end{tabular}
}
\end{center}
\caption{Comparison of edge features using SRNN-MaxNode.}
\label{tab:results4}
\end{table*}

\section{Conclusions}
\label{sec:summary}

In this work, we have proposed two variants of structural recurrent neural networks (SRNN) to recognize the actions of individuals as well as the activity of the entire group jointly. The advantage of the SRNN approach is that it explicitly models relations between individuals and all RNNs can be trained together using a single loss function. We evaluated the models on the Volleyball Dataset and showed that the SRNN model outperforms hierarchical LSTMs. 


\section*{Acknowledgment} The work has been financially supported by the ERC Starting Grant ARCA (677650). 
 
{\small
\bibliographystyle{ieee}
\bibliography{egbib}
}
\end{document}